%% file: main.tex
\documentclass[10pt,twocolumn,letterpaper]{article}

\usepackage{cvpr}              
\usepackage{multirow} 

\input{preamble}

\definecolor{cvprblue}{rgb}{0.21,0.49,0.74}
\usepackage[pagebackref,breaklinks,colorlinks,citecolor=cvprblue]{hyperref}

\def\paperID{5194} 
\def\confName{CVPR}
\def\confYear{2024}

\title{Surface Normal Estimation with Transformers}

\author{%
Barry Shichen Hu$^1$$^*$\quad
Siyun Liang$^1$$^*$\quad
Johannes Paetzold$^1$\quad \\
Huy H. Nguyen$^2\quad $
Isao Echizen$^{2,3}$\quad
Jiapeng Tang$^1$\quad
\vspace{0.2cm}\\
$^1$Technical University of Munich\quad \\
$^2$National Institute of Informatics, Japan\ \ \ \ \ \ \ \ 
$^3$University of Tokyo, Japan
}

\begin{document}

\twocolumn[{%
	\renewcommand\twocolumn[1][]{#1}%
	\maketitle
	\begin{center}
		\vspace{-0.35cm}
	\includegraphics[width=\linewidth]{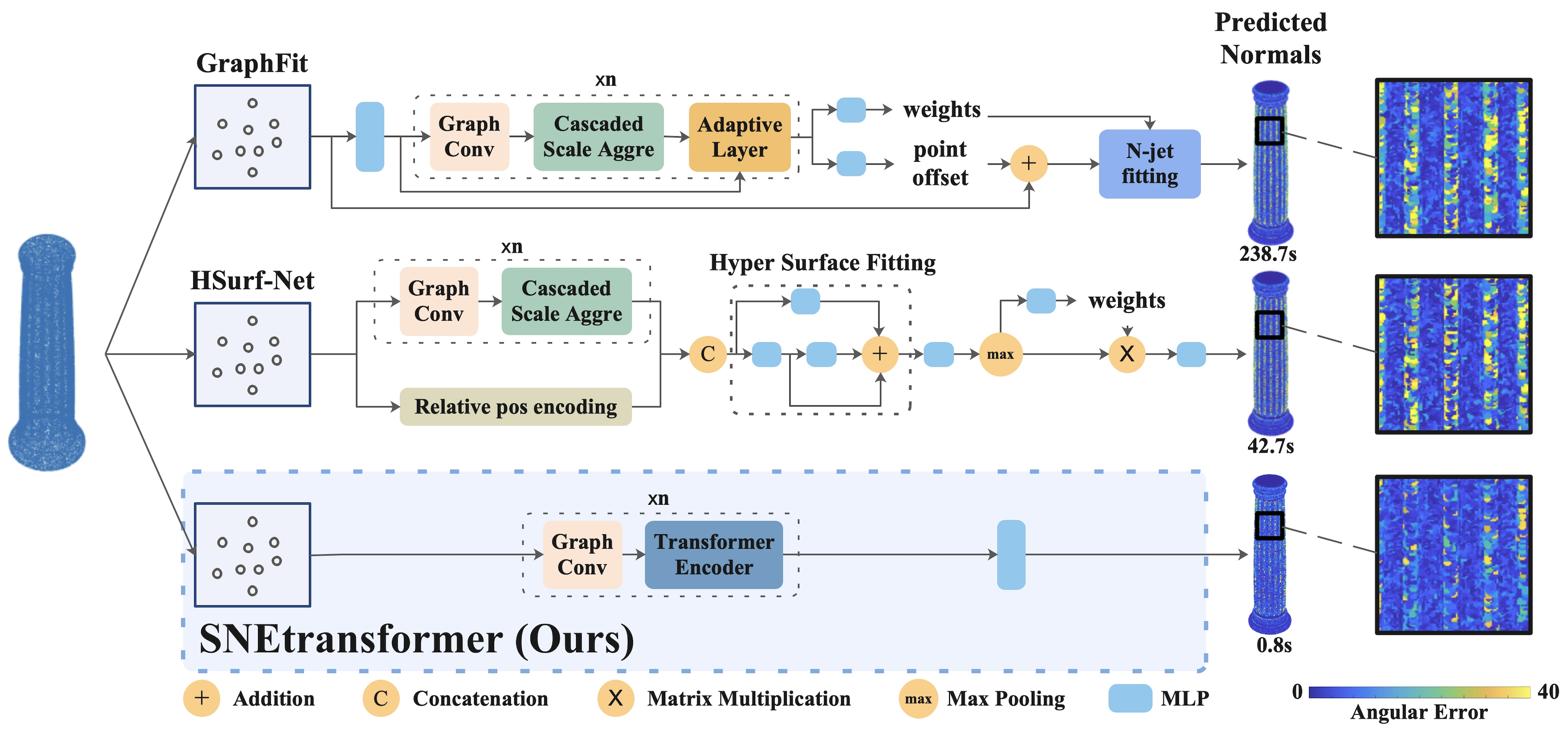}
		\captionof{figure}{
		We unify and simplify existing learning-based methods for surface normal estimation by proposing a straightforward Transformer-based model that directly predicts normals without relying on surface fitting. Our greatly simplified method not only achieves state-of-the-art performance but also exhibits significantly faster inference speed than previous works. In the figure, we present the simplified pipelines of existing works for comparison, and visualize the prediction error using a heat map. Inference times are recorded as well.
		}
		\vspace{-0.05cm}
		\label{fig:teaser}
	\end{center}
}]

\maketitle
\def\thefootnote{*}\footnotetext{Equal contribution.}

\begin{abstract}
We propose the use of a Transformer to accurately predict normals from point clouds with noise and density variations. Previous learning-based methods utilize PointNet variants to explicitly extract multi-scale features at different input scales, then focus on a surface fitting method by which local point cloud neighborhoods are fitted to a geometric surface approximated by either a polynomial function or a multi-layer perceptron (MLP). However, fitting surfaces to fixed-order polynomial functions can suffer from overfitting or underfitting, and learning MLP-represented hyper-surfaces requires pre-generated per-point weights. To avoid these limitations, we first unify the design choices in previous works and then propose a simplified Transformer-based model to extract richer and more robust geometric features for the surface normal estimation task. Through extensive experiments, we demonstrate that our Transformer-based method achieves state-of-the-art performance on both the synthetic shape dataset PCPNet, and the real-world indoor scene dataset SceneNN, exhibiting more noise-resilient behavior and significantly faster inference. Most importantly, we demonstrate that the sophisticated hand-designed modules in existing works are not necessary to excel at the task of surface normal estimation. The code, data, and pre-trained models are publicly available in \textcolor{red}{\href{https://anonymous.4open.science/r/E34CYRW-17E7}{https://anonymous.4open.science/r/E34CYRW-17E7}}.
\end{abstract}

  
\section{Introduction}
\label{sec:intro}

Estimating surface normals of point clouds is a fundamental problem in 3D computer vision that has a wide variety of downstream applications, such as point cloud denoising~\cite{lu2020deep, avron2010L1,sun2015denoising,lu2020low}, rendering~\cite{blinn1978simulation,gouraud1971continuous,phong1975illumination}, and reconstruction~\cite{kazhdan2006poisson,Fleishman_least_square}. While a significant amount of research has been dedicated to this topic, the accurate prediction of point cloud normals amid various types of noise, missing structures, and density variations remains a persistent challenge.

Existing methods address the surface normal estimation problem through either traditional surface fitting methods or more recent learning-based approaches. Traditional methods involve fitting planes or polynomials to a local neighborhood and then computing the normal from the estimated surface~\cite{hoppe1992surface, alexa2001point, huang2009consolidation,lange2005anisotropic, stewart1993early}. However, explicit surface fitting is sensitive to noise and outliers. Furthermore, it heavily relies on hand-tuned parameters, such as the order of the polynomial function, which can lead to underfitting or overfitting~\cite{, mitra2003estimating, pauly2002efficient, boulch2012fast, guennebaud2007algebraic}. In contrast, earlier learning-based methods like ~\cite{boulch2016deep, guerrero2018pcpnet, ben2019nesti, zhou2020normal, zhou2022refine} apply neural networks to directly regress the surface normal, thus bypassing explicit surface fitting and its associated challenges.

However, recent learning-based methods ~\cite{ben2020deepfit,zhu2021adafit, lenssen2020deep, graphfit, zhang2022geometry} have renewed interest in the traditional surface fitting paradigm, demonstrating that the integration of a neural network into this conventional approach led to superior performance compared to direct regression. These methods initially use a neural network, such as the PointNet family ~\cite{qi2017pointnet, qi2017pointnet++}, to learn point-wise weights of a neighborhood and then apply a classic geometric surface fitting algorithm, like n-jet fitting, to compute normals ~\cite{CAZALS2005121}. Following the idea of surface fitting, ~\cite{hsurf} innovatively proposes hyper-surface fitting by learning a set of MLP layers whose parameters interpret the geometric structures of a hyper-surface. While avoiding the model fitting problem associated with surface fitting methods, ~\cite{hsurf} relies on a set of handcrafted per-point weights that may not accurately reflect the true contribution of points to the surface fitting. To address these issues, ~\cite{NeAF} learns an angular field that points toward the ground truth normal, instead of directly predicting the surface normal. This method, however, requires extensive sampling and time-consuming optimization during testing. To mitigate the pitfalls of existing methods, we take a step back and ask the challenging question: 

\textit{Can rich geometric features be extracted directly from raw point clouds for normal estimation without relying on any handcrafted features or hand-designed modules?}

To address this question, we first analyze current learning-based methods for surface normal estimation and discover that, despite variations in network design, the fundamental design choices in existing works are centered around Graph Convolution, which preserves locality, and multi-scale feature fusion, which aggregates geometric features from larger to smaller scales. Therefore, in this work, we continue to use Graph Convolution for local neighborhood aggregation and explore the optimal features for the Graph operation. Additionally, we propose using a Transformer as an alternative for multi-scale feature extraction, contending that the Transformer can extract richer multi-scale features due to its superior capacity for modeling relationships and its expansive receptive field.

As a result, we propose \textbf{SNEtransformer}, a simplified and unified Transformer-based backbone that learns directly from point clouds for normal estimation. Experiments on synthetic and RGB-D scan datasets demonstrate that our backbone not only achieves state-of-the-art performance but also proves to be faster in inference and more resilient to noise compared to existing methods. In summary, our main contributions are:

\begin{itemize}
    \item We unify previous learning-based methods and propose the first Transformer-based model for end-to-end normal estimation without additional surface fitting steps.
    \item We demonstrate that our method achieves state-of-the-art accuracy and inference speed, showing greater resilience to noise in both synthetic and real-world scan datasets.
    \item Through comprehensive ablation studies, we identify the best design decisions that lead to increased accuracy.
\end{itemize}


\section{Related Work}


\paragraph{Learning-based Direct Regression Methods.}
Initial methods have been proposed to directly regress normal vectors from raw point clouds using neural networks. PCPNet~\cite{guerrero2018pcpnet} applies the PointNet architecture~\cite{qi2017pointnet} in multi-scale neighborhoods to extract geometric features based on which normals and curvatures of point clouds are estimated. Nesti-Net~\cite{ben2019nesti} follows a structure similar to PCPNet but proposes training multiple backbones on neighborhoods of different sizes, then uses a mixture-of-experts architecture~\cite{jacobs1991adaptive} to select the optimal backbone to predict the normal. Refine-Net~\cite{zhou2022refine} follows a two-stage design where it first computes an initial normal estimate and then deploys a deep neural network for refinement. Despite the simplicity of existing direct regression methods, they have demonstrated weaker performance in normal estimation tasks.

\vspace{-0.35cm}
\paragraph{Learning-based Surface Fitting Methods.}
Recent methods combine traditional surface fitting techniques with deep learning to achieve higher accuracy. DeepFit~\cite{ben2020deepfit} and AdaFit~\cite{zhu2021adafit} both use PointNet-based models~\cite{qi2017pointnet} to predict point-wise weights in a local patch and then apply n-jet fitting to estimate the normal~\cite{CAZALS2005121}. However, they suffer from underfitting or overfitting due to the fixed order of the polynomial function. Hsurf~\cite{hsurf} explores geometric priors in high-dimensional space but requires training with hand-crafted per-point weights. NeAF~\cite{NeAF} learns an angular field and applies extensive sampling and test time optimization to obtain surface normal vectors. Each of the aforementioned methods features hand-designed modules and comes with its own limitations. Instead, we propose a simpler yet more effective architecture that directly predicts surface normals from raw point clouds.

\vspace{-0.35cm}
\paragraph{Transformers in 3D Applications.} Transformers have gained increasing popularity since their introduction~\cite{vaswani2017attention}. Though they were originally introduced as a language model, they have proven to be effective in computer vision tasks~\cite{dosovitskiy2021image, redmon2015unified, chen2022unit3d, 3DVG_Transformer, zhu2021deformable}. Furthermore, Transformers serve as robust backbones for 3D applications as well. \citet{zhao2021point} proposed the use of a Transformer for point cloud classification and segmentation tasks. \citet{misra2021-3detr} applied a Transformer to 3D object detection without relying on predetermined query points. \citet{yu2021pointr} achieved state-of-the-art performance in 3D point cloud completion, while \citet{shit2022relationformer} utilized the Transformer's relational modeling power in 3D graph generation. Although Transformers have been shown to be effective for 3D tasks, no previous work has applied Transformers to surface normal estimation tasks. Therefore, we decided to explore this area by experimenting with the straightforward use of Transformers for point cloud normal estimation tasks.


\section{Method}
\label{sec:method}

\subsection{Preliminaries}
\label{sec:preliminaries}

\paragraph{Surface Normal Estimation.} Given a local point set centered at a query point \(\mathbf{p}\) as \(\mathcal{P} = \{\mathbf{p}_i \mid i = 1, \ldots, m\}\), the learning objective is to estimate the unoriented normal \(\mathbf{n_p}\) of the point \(\mathbf{p}\).

\noindent
\vspace{-0.35cm}
\paragraph{Graph Convolution.} Graph Convolution learns local geometric structures of a point cloud by first constructing a local neighborhood graph through the $k$-nearest neighbor algorithm centered at a query point~\cite{DGCNN}. We represent the resulting point cloud patch as coordinates \(\mathcal{X} = \{\mathbf{x}_1, \ldots, \mathbf{x}_k\} \subseteq \mathbb{R}^3\), their features \(\mathcal{F} = \{\mathbf{f}_1, \ldots, \mathbf{f}_k\} \subseteq \mathbb{R}^F\), and the graph as \(\mathcal{G} = (\mathcal{V}, \mathcal{E})\), where \(\mathcal{V} = \{1, \ldots, k\}\) and \(\mathcal{E} \subseteq \mathcal{V} \times \mathcal{V}\) are the nodes and edges, respectively.

\begin{figure*}[h]
\centering
\includegraphics[scale=0.3]{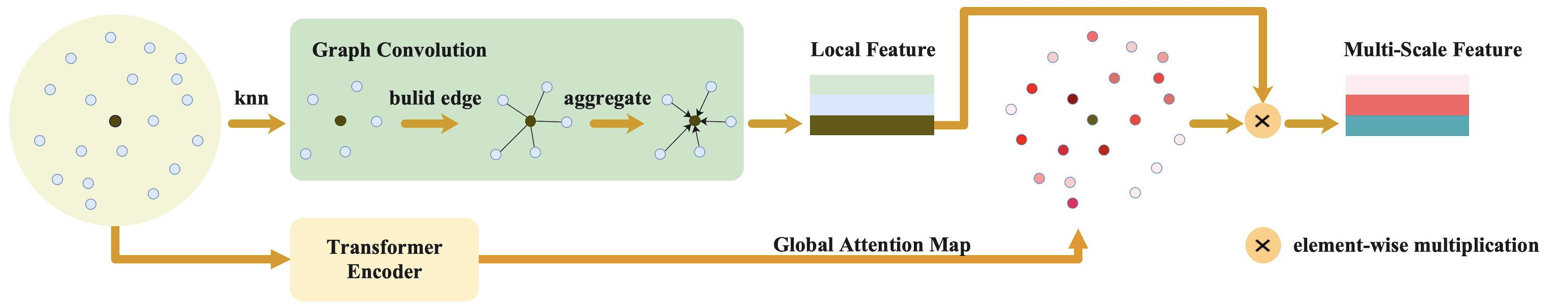}
\caption{Graph Convolution preserves locality, while the Transformer Encoder extracts multi-scale features. The global attention map assigns larger weights to `more reliable' points and smaller weights to `unreliable' ones, thereby functioning as a denoising mechanism.}
\label{fig:arch}
\end{figure*}

Then, one convolution step calculates and aggregates the edge features, as graphically illustrated in Figure \ref{fig:arch}, and the mathematical formula for the convolution operation is:

\begin{equation}
    \mathbf{f}_i^{\prime} = \square_{j : (i, j) \in \mathcal{E}} h_{\Theta}(\mathbf{f}_i, \mathbf{f}_j)
\end{equation}

\noindent
where \(\mathbf{f}_i\) is the feature vector of the point \(\mathbf{x}_i\), and \(\{\mathbf{f}_j : (i, j) \in \mathcal{E}\}\) are features of the nearest neighbors of \(\mathbf{x}_i\). The function \(h_{\Theta}: \mathbb{R}^F \times \mathbb{R}^F \rightarrow \mathbb{R}^{F'}\) is a learnable function parameterized by \(\Theta\) that extracts edge features, and \(\square\) is a symmetric aggregation function.

\vspace{-0.35cm}
\paragraph{Cascaded Scale Aggregation.} To explicitly extract multi-scale features from a point cloud patch, Zhu et al.~\cite{zhu2021adafit} proposed the Cascaded Scale Aggregation (CSA), which was later adopted by subsequent studies~\cite{graphfit, hsurf}. Essentially, CSA utilizes features from a larger scale to assist feature extraction at a smaller scale. For a point \( \mathbf{p} \), the scale \( s \) is defined as the size of the nearest neighbor point set \( \mathcal{N}_s(\mathbf{p}) \). A CSA layer considers two scales, \( s_k \) and \( s_{k+1} \), where \( s_{k+1} < s_k \) and \( \mathcal{N}_{s_{k+1}} \subseteq \mathcal{N}_{s_k} \). For a point \( \mathbf{p}\), we integrate its feature \( \mathbf{f}_{k} \) at scale \( s_k \) into its feature aggregation at scale \( s_{k+1} \), as follows:

\begin{equation}
    \mathbf{f}_{k+1}=\phi_k\left(\varphi_k\left(\operatorname{MaxPool}\left\{\mathbf{f}_{k, j} \mid \mathbf{p}_j \in \mathcal{N}_{s_k}\right\}\right), \mathbf{f}_{k}\right)
\end{equation}
\noindent
where both \(\phi_k\) and \(\varphi_k\) are Multi-layer Perceptrons (MLPs). The motivation is that the larger scale provides information about broader surfaces, while the smaller scale includes more detailed local features for surface fitting~\cite{zhu2021adafit}. In subsequent sections, we demonstrate that the Transformer is a superior method for multi-scale feature extraction compared to CSA.

\vspace{-0.35cm}
\paragraph{Attention Mechanism.} Attention models the relations between inputs. The inputs consist of queries, keys, and values, where the queries and keys are of the same dimension $d_k$, and values are of dimension $d_v$. To compute the attention scores,~\cite{vaswani2017attention} computes the dot products of all queries with all keys, divided by $\sqrt{d_k}$, and then applies a softmax function. The resulting mathematical formulation is:

\begin{equation}
    \operatorname{Attention}(\mathbf{Q}, \mathbf{K}, \mathbf{V})=\operatorname{softmax}\left(\frac{\mathbf{Q} \mathbf{K}^T}{\sqrt{d_k}}\right) \mathbf{V}    
\end{equation}
\noindent
where $\mathbf{Q}$, $\mathbf{K}$, and $\mathbf{V}$ are the query, key and value matrices~\cite{vaswani2017attention}. Self-attention simply means that the $\mathbf{Q}$, $\mathbf{K}$, and $\mathbf{V}$ are derived from the same input features, and the Transformer Encoder architecture utilizes the self-attention mechanism to extract the relations between the inputs. 

We apply Transformer Encoder layers to extract multi-scale geometry for two reasons. First, the attention mechanism `attends' to points in both smaller and larger neighborhoods, thus naturally extracting multi-scale features. Second, attention models the contribution of each input point to the calculation of the normal vector, and such contributions are modeled by the attention scores. This serves as a denoising mechanism, particularly useful in the case of noisy input where the attention learns to weigh the importance of input points instead of indiscriminately favoring smaller neighborhoods, as CSA does. Therefore, we hypothesize that Transformers lead to more noise-agnostic behavior. This hypothesis is verified in Section \ref{sec:results}, and Figure \ref{fig:visualization} visualizes the weights predicted by CSA and Transformer.


\subsection{Analysis of Alternative Methods}
We analyze recent state-of-the-art models to identify design patterns that lead to improved surface normal estimation performance. The architectures of HSurf-Net~\cite{hsurf}and GraphFit~\cite{graphfit} are documented in Figure \ref{fig:teaser}.
\label{sec:previous methods}

\vspace{-0.35cm}
\paragraph{AdaFit.} The backbone of AdaFit is based on the Cascaded Scale Aggregation (CSA) module. It employs a series of CSA and MLP layers to aggregate geometric features from broader neighborhoods down to the smallest neighborhood~\cite{zhu2021adafit}. Then, it deploys two MLP heads to predict the point-wise offsets and weights. The point-wise offsets are used to denoise the input point cloud, and then the weights are applied for n-jet fitting on the denoised point patch to predict the normal vector at the query point ~\cite{zhu2021adafit}.

\vspace{-0.35cm}
\paragraph{GraphFit.} GraphFit consists of a series of CSA-based Graph Convolution and adaptive layers~\cite{graphfit}. Given the feature vector set $\mathcal{F} = \left\{\mathbf{f}_i \mid i=1,2, \ldots, m\right\}$ of a local point set $\mathcal{P}$, the Graph Convolution is formulated as follows:

\begin{equation}
    \mathbf{f}_i^{\prime}=\max _{j \in \mathcal{N}(i)} \phi\left( \left[\mathbf{f}_j-\mathbf{f}_i, \mathbf{f}_i\right]\right)
\end{equation}
where $\mathbf{f}_i$ is the feature at point $\mathbf{p}_i$, and $\{\mathbf{f}_j \mid j \in \mathcal{N}(i)\}$ is the set of features of point $\mathbf{p}_i$'s neighbours. $[\cdot, \cdot]$ is the concatenation operation and $\phi$ represents an MLP~\cite{graphfit}.
Then, CSA is used to include features obtained from current Graph Convolution into the next level.
Following one block of Graph Convolutions and CSA, which outputs the aggregated feature set $\mathcal{F}^{\prime} = \left\{\mathbf{f}_i^{\prime} \mid i=1,2, \ldots, m\right\}$, GraphFit adaptively updates the per-point feature as follows:

\begin{align}
    \overline{\mathcal{F}} &= \mathbf{s}_{\mathcal{F}} \odot \mathcal{F} + \left(1-\mathbf{s}_{\mathcal{F}}\right) \odot \mathcal{F}^{\prime}
\end{align}

\noindent
where $\mathbf{s}_{\mathcal{F}}$ represents element-wise weights predicted from $\mathcal{F}$ and $\mathcal{F}^{\prime}$ by an MLP. Similar to AdaFit, extracted features are used to predict the per-point offsets and weights for n-jet surface fitting \cite{graphfit}.

\vspace{-0.35cm}
\paragraph{HSurf-Net.} Like~\cite{graphfit, zhu2021adafit}, HSurf-Net utilizes Graph Convolution based operation to extract local features and CSA variant to fuse features from larger to smaller scales~\cite{hsurf}. However, to prevent overfitting or underfitting due to the fixed order of the polynomial function, HSurf-Net proposes the use of an MLP to represent a hypersurface. To conduct hypersurface fitting, HSurf-Net predicts a set of per-point weights and element-wise multiplies these weights with the point features extracted. The resulting feature set is then fed into a block of MLP and pooling layers to predict the normal. To guide the model in predicting the correct weight for each point, HSurf-Net uses pre-generated target weights following the method in \cite{ZHANG2022103119}.



\vspace{-0.35cm}
\paragraph{Summary on Common Design Patterns.} Despite differences in surface fitting and specific hand-crafted network modules, the common designs of existing methods include Graph Convolution and multi-scale feature fusion with CSA. Thus, we propose a simple Transformer-based backbone that unifies existing works.

\subsection{Proposed Architecture}
\label{sec:architecture}
Our backbone consists of multiple layers of enhanced Graph Convolution \cite{DGCNN} and Transformer Encoder \cite{vaswani2017attention}, as illustrated in Fig. \ref{fig:arch}. At each layer, we first update each point's feature through a Graph Convolution operation. Then, the point features are directly fed as input to a Transformer Encoder layer.

\subsubsection{Enhanced Graph Convolution}
\label{sec:graph_conv}
Inspired by the relative position encoding discussed in \cite{hsurf} and the graph convolution presented in \cite{graphfit}, we propose an enhanced convolutional approach for feature aggregation within a local neighborhood. Consider a local point cloud obtained by executing the $k$-nearest neighbors algorithm centered at a point with coordinates $\mathbf{x}_c$, resulting in a graph-structured point set represented by Cartesian coordinates $\left\{\mathbf{x}_i \mid i=1,2, \ldots, k\right\} \in \mathbb{R}^{k \times 3}$ and their corresponding features $\left\{\mathbf{f}_i \mid i=1,2, \ldots, k\right\} \in \mathbb{R}^{k \times F}$. To aggregate the features from the local neighbourhood to $\mathbf{x}_c$, we first construct the edge features between $\mathbf{x}_c$ and $\mathbf{x}_j$ as follows:
\begin{equation}
    \mathbf{e}_{c j}=\phi\left([\mathbf{x}_j - \mathbf{x}_c, \mathbf{x}_c, \mathbf{x}_j, \mathbf{f}_j, \mathbf{f}_j - \mathbf{f}_c]\right), j \in \mathcal{N}(c)
\end{equation}
where $[\cdot,\cdot]$ represents the concatenation operation, and $\phi$ is implemented as an MLP. We then output the local neighborhood information for $\mathbf{x}_c$:
\begin{equation}
    \mathbf{f}_c^{\prime}=\max _{j \in \mathcal{N}(c)} \mathbf{e}_{c j}
\end{equation}
Our graph convolution operation not only aggregates local features to preserve locality, but also encodes positional information and edge features for use in the subsequent Transformer Encoder layer.

\subsubsection{Transformer Layer}
\label{sec:trans_layer}
We utilize the Transformer Encoder Layer, as proposed in \cite{vaswani2017attention}, to extract multi-scale geometric features. The architecture is depicted in Figure \ref{fig:transformer_arch} in the supplementary material. Specifically, features extracted from the graph convolution are directly fed into a Transformer Encoder layer. Instead of limiting the attention operation to a local neighborhood of a point, global attention is employed among all points in the input. The experimental results, detailed in Section \ref{sec:Ablation}, demonstrate that this global attention mechanism leads to improved outcomes.

\subsubsection{Loss Function}
\label{sec:loss_function}
Our goal is to predict the unoriented normal vector; hence, we apply the sin loss between the predicted normal vectors of the point cloud patch, $\hat{\mathbf{n}}_\mathbf{p}$, and the ground truth, $\mathbf{n}_\mathbf{p}$:

\begin{equation}
    L=\left\|\hat{\mathbf{n}}_\mathbf{p} \times \mathbf{n}_\mathbf{p}\right\|
\end{equation}

\subsubsection{Comparison to Alternative Backbones}
\label{sec:sneregression}
Our model architecture is greatly simplified compared to alternative methods. First, we do not rely on a surface fitting scheme like those described in \cite{hsurf, zhu2021adafit, graphfit, ben2020deepfit}; instead, we directly predict the normal vectors for the entire input point cloud patch. Second, we do not explicitly extract multi-scale features by operating the backbone at different scales of the point cloud; rather, we leverage the Transformer's ability to model relationships and implicitly extract multi-scale features. Third, instead of using carefully hand-designed modules to extract geometric features from the input, we apply simple Graph Convolution layers and Transformer encoder layers. We demonstrate that a straightforward combination of Graph Convolution with a Transformer is sufficient to accurately predict normal vectors.

\section{Results}
\label{sec:results}

We first explain the experimental setup, then demonstrate the quality of normal estimation on the widely-used synthetic dataset PCPNet \cite{guerrero2018pcpnet}, the real-world scan dataset SceneNN and Semantic3D \cite{scenenn-3dv16, hackel2017isprs}. Finally, we use ablation studies to identify the design choices that enable our method to be more accurate. For additional qualitative visualizations, please refer to the supplementary material.

\subsection{Experimental Setup}
\label{sec:setup}
\paragraph{Data Preprocessing.}
Similar to ~\cite{hsurf, graphfit, ben2020deepfit}, SNEtransformer takes in a local patch of 700 points obtained by the $k$-nearest neighbors algorithm at a query point. Following ~\cite{hsurf, graphfit, guerrero2018pcpnet}, to remove unnecessary degrees of freedom, we normalize each point's coordinates by the patch radius and rotate the points into a coordinate system defined by Principal Component Analysis~\cite{hsurf}. Given a point cloud patch, instead of only predicting the normal vector of a query point as in ~\cite{hsurf, graphfit, guerrero2018pcpnet}, SNEtransformer predicts the normal vectors of the entire query point neighborhood.

\paragraph{Evaluation Metrics.} 
\label{sec:eval_metrics}
We adopt the angular Root Mean Squared Error (RMSE) between the predicted normal and the ground truth to evaluate the estimation results~\cite{guerrero2018pcpnet}. Suppose a point cloud $\mathcal{P}$ 's predicted normal set is $\hat{\mathcal{N}}(\mathcal{P})=\left\{\hat{\mathbf{n}}_i \in \mathbb{R}^3\right\}_{i=1}^{m}$, and ground truth normal set is $\mathcal{N}(\mathcal{P})=\left\{\mathbf{n}_i \in \mathbb{R}^3\right\}_{i=1}^{m}$. The RMSE error is calculated as:
\begin{equation}
    \operatorname{RMSE}(\hat{\mathcal{N}}(\mathcal{P}))=\sqrt{\frac{1}{m} \sum_{i=1}^{m} \arccos ^2\left(\hat{\mathbf{n}}_i, \mathbf{n}_i\right)}
\end{equation}

\noindent
Following \cite{hsurf, zhu2021adafit, guerrero2018pcpnet}, we also use  the metric of the percentage of good points PGP($\alpha$) to analyze the error distribution of the predicted normal:
\begin{equation}
    \operatorname{PGP}(\alpha)=\frac{1}{m} \sum_{i=1}^{m} I\left(\arccos \left(\hat{\mathbf{n}}_i, \mathbf{n}_i\right)<\alpha\right), \alpha \in \left[0^\circ,30^\circ\right]
\end{equation}
\noindent 
where $I$ is the indicator function. PGP($\alpha$) measures the percentage of normal predictions with errors that fall below various angle thresholds denoted by $\alpha$.

\vspace{-0.35cm}
\paragraph{Implementation Details.} 
\label{sec:implementation}
For training, we use an Adam optimizer with a learning rate of $2\times10^{-4}$ and a batch size of 32. The learning rate is decreased by a factor of 0.005 every epoch. Our method is trained for 250 epochs, during which we randomly sample 100,000 point patches from the training set in each epoch. Experiments are conducted on a cluster of NVIDIA A100 GPUs.
\begin{figure*}
	\centering
	\includegraphics[width=0.88\linewidth]{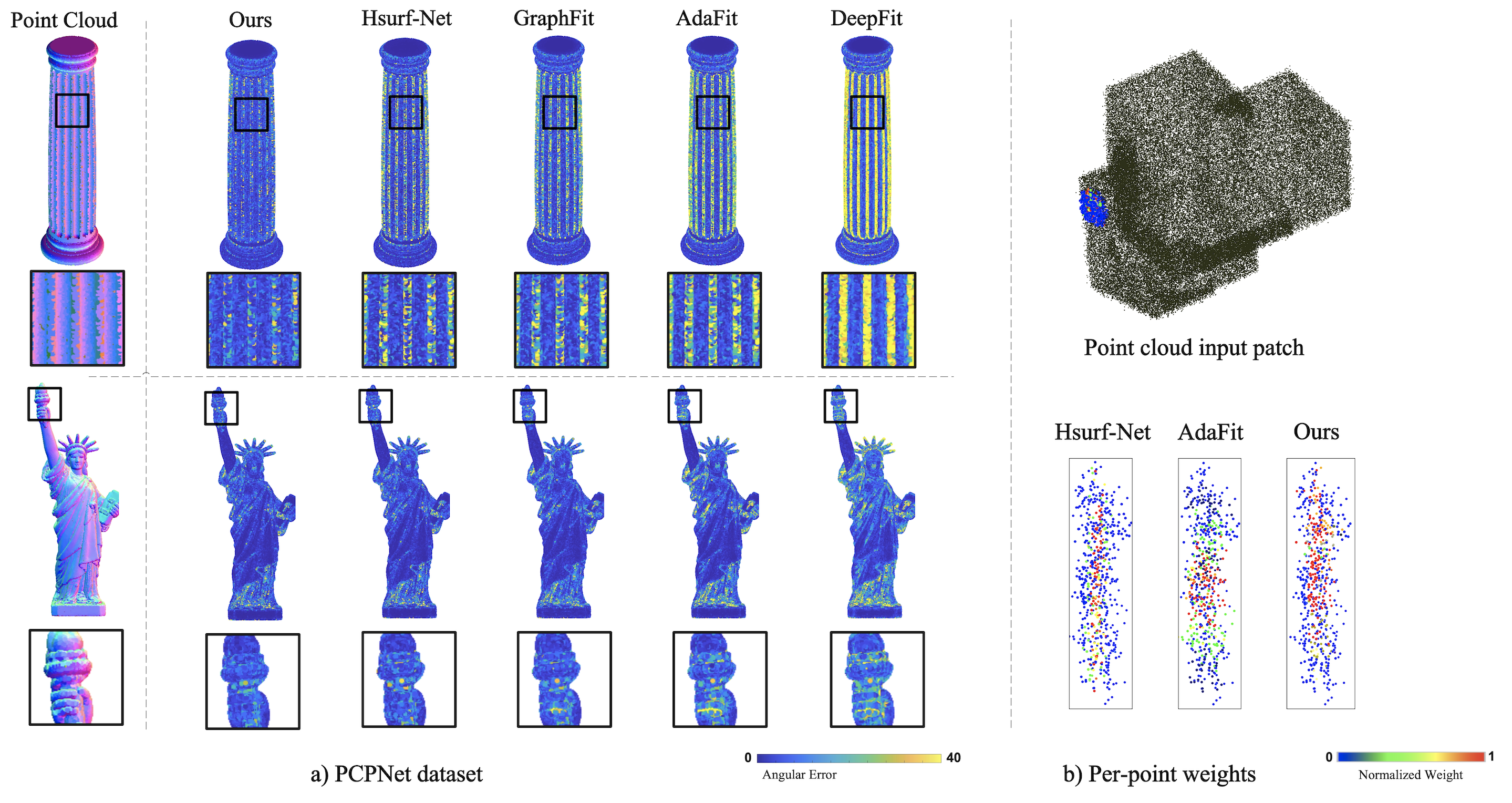}
    \vspace{-0.15cm}
	\caption{a) Qualitative results on PCPNet dataset. The point cloud heatmap reflects the error on the normal estimation. b) Visualization of the per-point weight. CSA (AdaFit) favors smaller neighborhoods indiscriminately, while HSurf-Net is trained with weights that prioritize `on surface' points. Meanwhile, the Transformer acquires optimal global attention weights through training on raw point cloud data.
	}
    \vspace{-0.15cm}
	\label{fig:visualization}
\end{figure*}
\input{tables/pcpnet_scenenn.tex}
\begin{figure*}
	\centering
	\includegraphics[width=\linewidth]{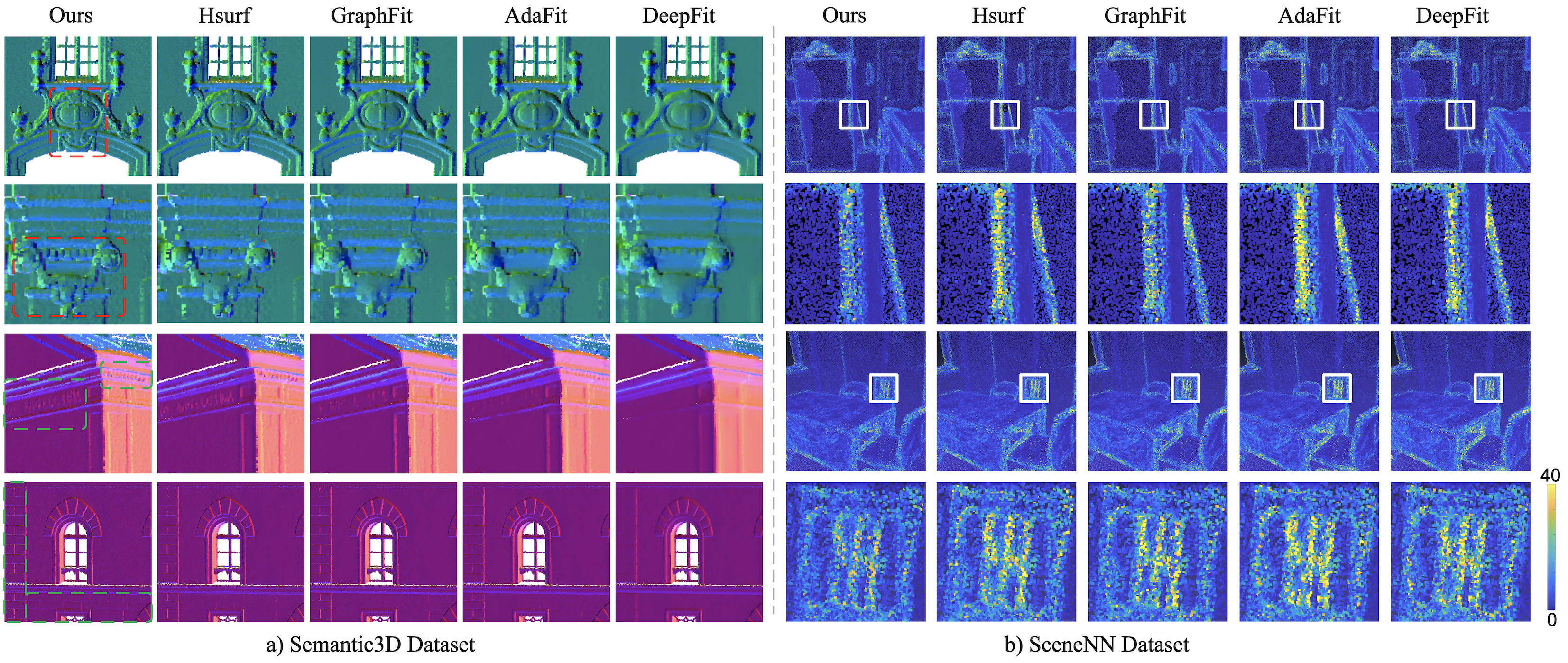}
	\caption{a) Visualization of predicted normals on the Semantic3D dataset. Our method preserves sharper geometric details, as highlighted by the red and green border regions. b) Error visualization of noisy point clouds in the SceneNN datasets. Point colors correspond to the angular error mapped onto a heatmap. SNEtransformer predicts more accurate normals than baselines when the input is affected by noise.
	}
    \vspace{-0.15cm}
	\label{fig:semantic3d_scenenn}
\end{figure*}
\begin{figure*}
	\centering
	\includegraphics[width=\linewidth]{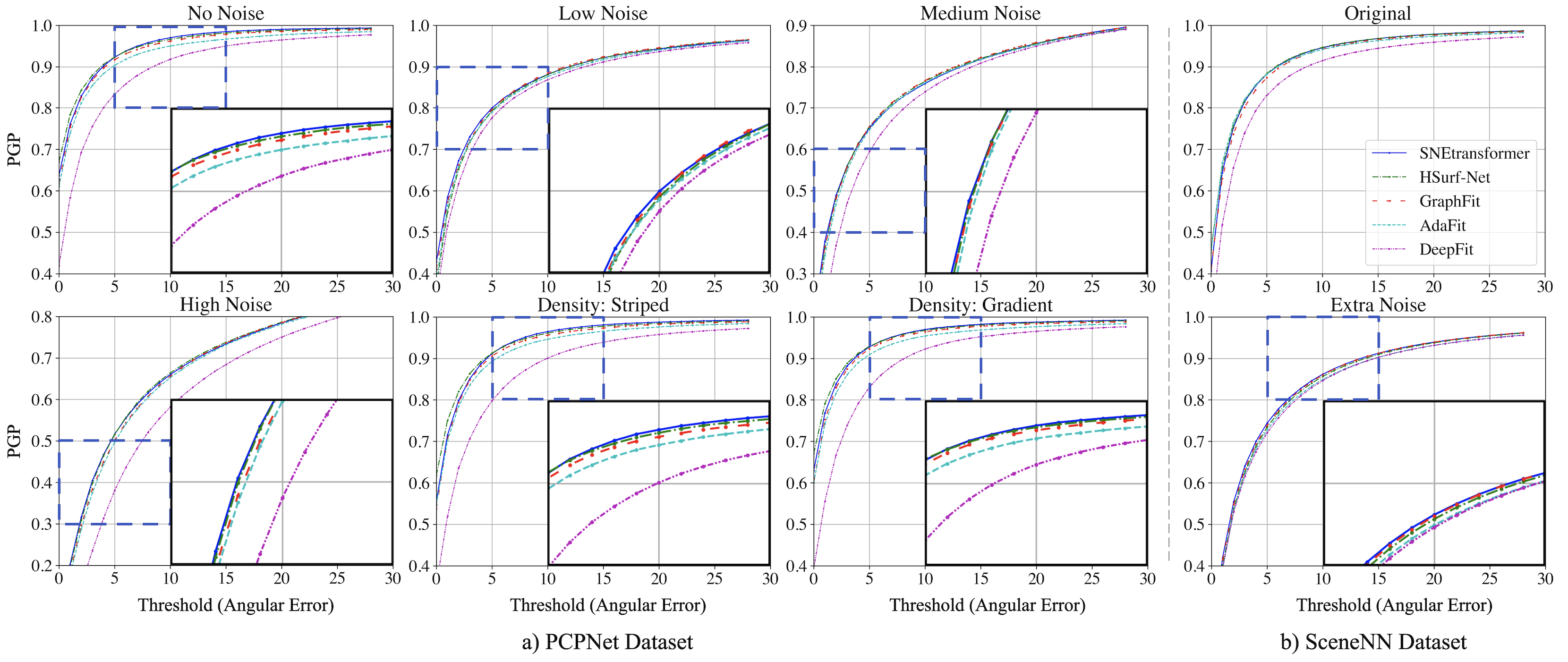}
	\vspace{-0.15cm}
	\caption{Percentage of Good Points (PGP) graphs for the PCPNet and SceneNN datasets. The area under the blue color is enlarged and displayed in a black pane. Our method
produces high-quality estimations in noisy settings.}
    \vspace{-0.35cm}
	\label{fig:pgp}
\end{figure*}

\subsection{Results on PCPNet}
\label{sec:PCPNet}
PCPNet is a point cloud normal estimation dataset comprising synthetic shapes and 3D scanned objects. The training set contains eight point clouds, and the test set consists of nineteen. Following \cite{hsurf, graphfit, zhu2021adafit}, SNEtransformer is trained on point clouds with various levels of Gaussian noise—none, low, medium, and high—and is evaluated against point clouds with different Gaussian noise levels, as well as two additional settings where the point density is inconsistent. The quantitative evaluation results, presented in Table~\ref{table:pcpnet_scenenn}, demonstrate that SNEtransformer outperforms existing methods in almost all scenarios. Figure~\ref{fig:pgp} displays the PGP curves under all noise conditions, and a visual comparison of the normal prediction error output by SNEtransformer and existing methods is shown in Figure~\ref{fig:visualization}. It is evident that our method produces more accurate normal estimations in various testing scenarios.

\subsection{Results on SceneNN}
\label{sec:SceneNN}
SceneNN is an RGB-D scan dataset captured in various indoor settings. Following \cite{hsurf}, we first train the SNEtransformer on the PCPNet dataset, then evaluate the trained model on SceneNN without fine-tuning to explore the model's scalability. Due to sensor errors, the data naturally contains noise, presenting a good opportunity to test the model's noise agnosticism. We use the same evaluation settings as \cite{hsurf} and report the numerical results in Table~\ref{table:pcpnet_scenenn} and the visual results in Figure~\ref{fig:visualization}. Table~\ref{table:pcpnet_scenenn} shows that SNEtransformer generalizes well to real-world data and outperforms previous methods on both original and extra-noise settings. Figure~\ref{fig:pgp} presents the PGP curves under original and extra-noise conditions, demonstrating that our method produces high-quality estimations for indoor scanning.

\subsection{Results on Semantic3D}
\label{sec:Semantic3D}
We visualize the normal estimation results on the outdoor scanning dataset Semantic3D in Figure~\ref{fig:semantic3d_scenenn}, despite the absence of ground truth data for normals. It is apparent that our method preserves finer details such as carved patterns on doors, grooves between bricks, and letters on buildings—details that other methods tend to oversmooth. This suggests that our method also provides higher-quality normal estimation in outdoor scanning scenarios.

\begin{figure*}
	\centering
	\includegraphics[width=0.78\linewidth]{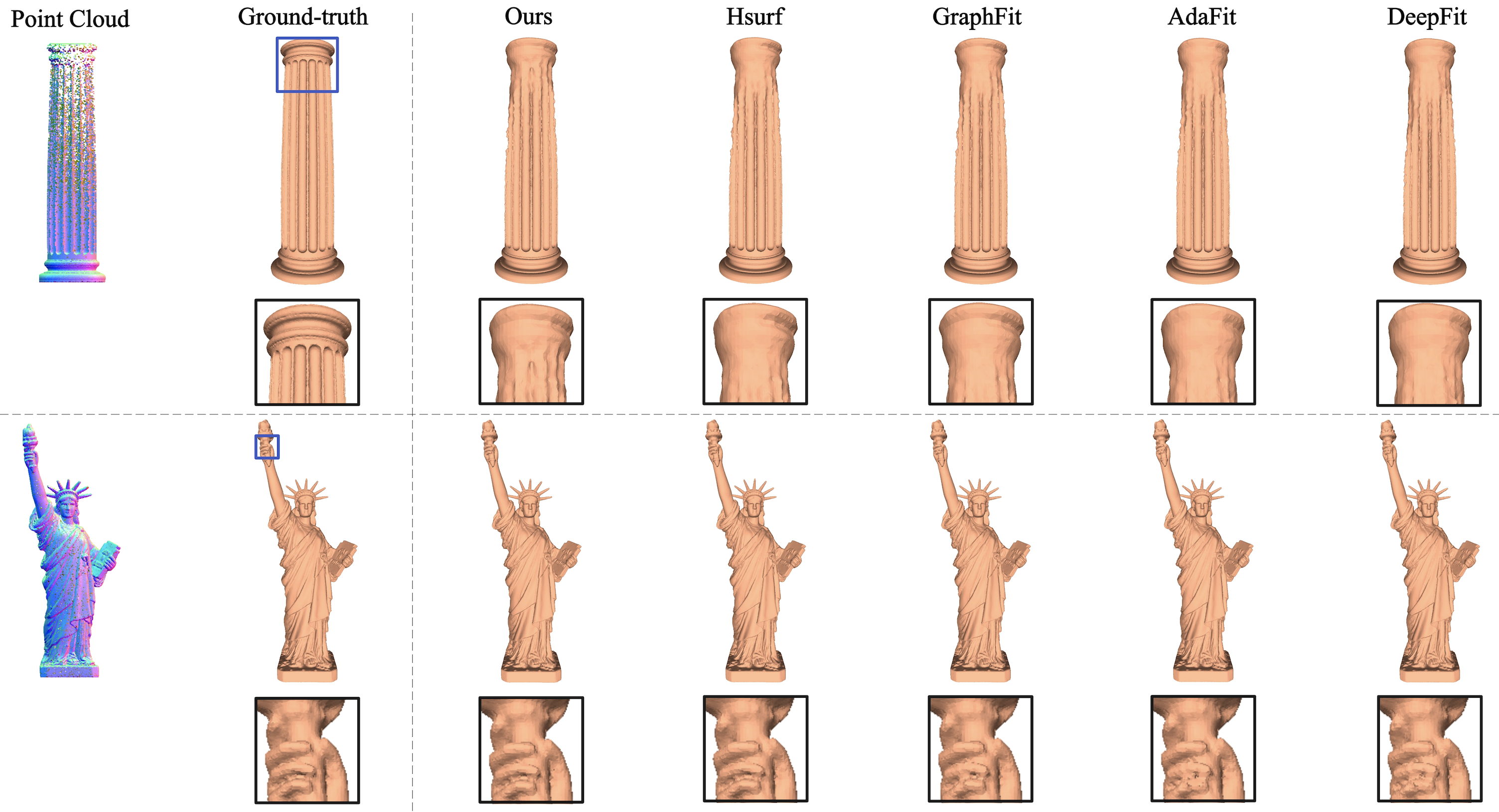}
    \vspace{-0.15cm}
	\caption{Poisson surface reconstruction on point clouds from the PCPNet dataset. Normals predicted by the SNEtransformer help recover finer details in point clouds with variant density noise (top) and Gaussian noise (bottom), as highlighted in the windows below each shape.}
    \vspace{-0.15cm}
	\label{fig:poisson}
\end{figure*}

\subsection{Application to Surface Reconstruction Task }
\label{sec:poisson}
We apply Poisson reconstruction~\cite{hoppe1992surface} on PCPNet point clouds with the per-point normal predicted by SNEtransformer. Figure~\ref{fig:poisson} shows that SNEtransformer helps recover finer details in areas with complex local geometry like the hand of the Statue of Liberty. Notably, our method enhances geometry reconstruction when the input is noisy.

\subsection{Ablation Studies}
\label{sec:Ablation}
\paragraph{Graph Convolution and Transformer.} 
To validate the effectiveness of Graph Convolution and the Transformer, we conduct ablation studies on them and report the results in Table~\ref{table:abla_arch}. We observed that the performance of normal estimation degrades when either Graph Convolution or the Transformer is removed from the network. However, the accuracy of normal estimation degrades even further when the Transformer architecture is removed, which demonstrates its significant effectiveness.

\vspace{-0.35cm}
\paragraph{Global Attention or Local Attention.}
To demonstrate the effectiveness of global attention, we compare its performance with that of local attention. To implement local attention, we first run the $k$-nearest neighbors algorithm at each point in the point cloud patch and then apply attention only within the set of nearest neighbor points. The results in Table~\ref{table:abla_arch} show that applying attention on a global scale improves the results and leads to noise-agnostic behavior. This validates our assumption that global attention allows the network to attend to any points it deems helpful for the estimation tasks, thereby increasing resilience to noise.
\input{tables/ablation_arch}
\input{tables/ablation_graph_feature}

\vspace{-0.35cm}
\paragraph{Ablation on Features for Graph Convolution.}
We have explored various features for inclusion in Graph Convolution, and the results are listed in Table~\ref{table:abla_graph_conv_features}. In summary, there are four main features: the 3D coordinates of the query point and its neighboring points (denoted as $xyz$), the difference in 3D coordinates between the query point and its neighbors (denoted as $\Delta_{xyz}$), the features of the query point and its neighboring points (denoted as $\mathbf{f}$), and the difference in features between the query point and the neighbors (denoted as $\Delta_\mathbf{f}$). We conclude that both $\Delta_{xyz}$ and $\Delta_{\mathbf{f}}$ contribute to better estimation results.


\section{Conclusion}
In this paper, we introduce the SNEtransformer, a Transformer-based model that accurately predicts surface normals. We demonstrate that a straightforward combination of Graph Convolution with a Transformer is sufficient to achieve state-of-the-art performance, without the need for hand-designed modules. Our model unifies existing approaches and is proven to be noise-agnostic, as evidenced by extensive experiments on both indoor and outdoor datasets. Lastly, we showcase the potential of our method in various downstream applications.

\section*{Acknowledgements}
This work was partially supported by JSPS KAKENHI Grant JP21H04907, and by JST CREST Grants JPMJCR18A6 and JPMJCR20D3, Japan.
{
    \small
    \bibliographystyle{ieeenat_fullname}
    \bibliography{main}
}
\newpage


\end{document}

%% file: preamble.tex
\usepackage[dvipsnames]{xcolor}


%% file: tables/pcpnet_scenenn.tex
\begin{table*}[t]
	\centering
	\footnotesize
	\setlength{\tabcolsep}{1.2mm}
	\begin{tabular}{|l|c|l|rrrr|rr|r|rr|r|}
		\toprule
		\multirow{3}{*}{Category} & \multirow{3}{*}{Year}& \multirow{3}{*}{Approach} & \multicolumn{7}{c|}{PCPNet Dataset} & \multicolumn{3}{c|}{SceneNN Dataset} \\
		\cline{4-13}
		& & & \multicolumn{4}{c}{Noise $\sigma$} & \multicolumn{2}{|c|}{Density} & \multirow{2}{*}{\textbf{Average}}    & Orig- & Extra & \multirow{2}{*}{\textbf{Average}}  \\
		& & & None & 0.12\% & 0.6\% & 1.2\%      & Stripes & Gradient            &  &  inal & Noise   \\
		\midrule
		PCA~\cite{hoppe1992surface} 	         & 1992 & Classical surface fitting & 12.29 & 12.87 & 18.38 & 27.52 & 13.66 & 12.81 &   16.25   & 15.93 & 16.32 &   16.12   \\
		Jet~\cite{cazals2005estimating}          & 2005 & Classical surface fitting & 12.35 & 12.84 & 18.33 & 27.68 & 13.39 & 13.13 &   16.29   & 15.17 & 15.59 &   15.38   \\
		HoughCNN~\cite{boulch2016deep}         & 2016 & Direct regression & 10.23 & 11.62 & 22.66 & 33.39 & 11.02 & 12.47 &   16.90   & - & - & - \\
		PCPNet~\cite{guerrero2018pcpnet}         & 2018 & Direct regression & 9.64 & 11.51 & 18.27 & 22.84 & 11.73 & 13.46 &    14.58   & 20.86 & 21.40 &   21.13   \\
		Nesti-Net~\cite{ben2019nesti}            & 2019 & Direct regression & 7.06 & 10.24 & 17.77 & 22.31 & 8.64 & 8.95 &      12.49   & 13.01 & 15.19 &   14.10   \\
		Lenssen \etal~\cite{lenssen2020deep}     & 2020 & Learning-based surface fitting & 6.72 & 9.95 & 17.18 & 21.96 & 7.73 & 7.51 &    11.84    & 10.24 & 13.00 &   11.62   \\
		DeepFit~\cite{ben2020deepfit}            & 2020 & Learning-based surface fitting & 6.51 & 9.21 & 16.73 & 23.12 & 7.92 & 7.31 &    11.80    & 10.33 & 13.07 &   11.70   \\
        Refine-Net~\cite{zhou2022refine}         & 2022 & Direct regression & 5.92 & 9.04 & 16.52 & 22.19 & 7.70 & 7.20 &    11.43    & 18.09 & 19.73 &   18.91   \\
        Zhang \etal~\cite{zhang2022geometry} & 2022 & Learning-based surface fitting & 5.65 & 9.19 & 16.78 & 22.93 & 6.68 & 6.29 &    11.25    & 9.31  & 13.11 &   11.21   \\
        Zhou \etal~\cite{zhou2021improvement}& 2021 & Learning-based surface fitting & 5.90 & 9.10 & 16.50 & 22.08 & 6.79 & 6.40 &    11.13    & - & - & - \\      
        AdaFit~\cite{zhu2021adafit}              & 2021 & Learning-based surface fitting  & 5.19 & 9.05 & 16.44 & 21.94 & 6.01 & 5.90 &    10.76    & 8.39  & 12.85 &   10.62   \\
        GraphFit ~\cite{graphfit}                & 2022 & Learning-based surface fitting  & 4.45 & \textbf{8.74} & 16.05 & 21.64 & 5.22 & 5.48 &    10.26    & 7.99  & 12.17 & 10.08            \\
        NeAF ~\cite{NeAF}                        & 2023 & Angular field & 4.20 & 9.25 & 16.35 & 21.74 & 4.89 & 4.88 & 10.22 & -  & - & -            \\
		HSurf-Net ~\cite{hsurf}                  & 2022 & Learning-based surface fitting & 4.17 & 8.78 & 16.25 & 21.61 & 4.98 & 4.86 &    10.11    & 7.55  & 12.23 & 9.89            \\
		SNEtransformer (Ours)  & 2023 & Direct regression & \textbf{3.99} & 8.97 & \textbf{15.85} & \textbf{20.98} & \textbf{4.81} & \textbf{4.67} &  \textbf{9.88} & \textbf{7.44} & \textbf{12.14} & \textbf{9.79}            \\
		SNEdiffusion (Ours) & 2023 & Diffusion & 4.00 & 8.88 & 16.25 & 21.37 & 4.96 & 4.89 & 10.05 & \textbf{-} & \textbf{-} & \textbf{-} \\
		SNEdiffusion as regression model(Ours) & 2023 & Direct regression & 3.93 & 8.90 & 16.27 & 21.40 & 4.82 & 4.64 & 10.00 & \textbf{-} & \textbf{-} & \textbf{-} \\
		\bottomrule
	\end{tabular} 
	\caption{Normal angle RMSE results on the PCPNet and SceneNN dataset,
		sorted by the values (lowers are better) on the PCPNet dataset. As a direct regression method, SNEtransformer outperforms existing learning-based surface fitting methods significantly in noisy scenarios.
    }
    \label{table:pcpnet_scenenn}
\end{table*}

%% file: tables/ablation_arch.tex
\begin{table}[t]
    \centering
    \footnotesize
    \setlength{\tabcolsep}{1.2mm}
    
    \begin{tabular}{l|c|ccc}
        \toprule
        \multirow{3}{*}{Noise level} & \multicolumn{4}{c}{PCPNet Dataset}  \\
        \cline{2-5}
        & \multirow{2}{*}{Ours} & Ours w/o & Ours w/o & Ours with   \\
        & &  Transformer & GC & local attention   \\
        \midrule
        None                    & 3.99  & 5.95 \tiny{\textbf{\textcolor{gray}{(+1.95)}}}& 5.05 \tiny{\textbf{\textcolor{gray}{(+1.05)}}}& 4.61 \tiny{\textbf{\textcolor{gray}{(+0.61)}}} \\
        $\sigma$=0.12\%         & 8.97  & 9.77 \tiny{\textbf{\textcolor{gray}{(+0.80)}}}& 9.53 \tiny{\textbf{\textcolor{gray}{(+0.56)}}}& 9.30 \tiny{\textbf{\textcolor{gray}{(+0.33)}}} \\
        $\sigma$=0.6\%          & 15.85 & 17.98 \tiny{\textbf{\textcolor{gray}{(+2.13)}}}& 17.01 \tiny{\textbf{\textcolor{gray}{(+1.16)}}}& 17.20 \tiny{\textbf{\textcolor{gray}{(+1.35)}}} \\
        $\sigma$=1.2\%          & 20.98 & 22.56 \tiny{\textbf{\textcolor{gray}{(+1.57)}}}& 22.28 \tiny{\textbf{\textcolor{gray}{(+1.29)}}}& 22.14 \tiny{\textbf{\textcolor{gray}{(+1.15)}}}  \\
        Density (stripes)       & 4.81  & 6.67 \tiny{\textbf{\textcolor{gray}{(+1.85)}}}& 6.08 \tiny{\textbf{\textcolor{gray}{(+1.26)}}}& 5.52 \tiny{\textbf{\textcolor{gray}{(+0.70)}}}      \\
        Density (gradients)     & 4.67  & 6.14 \tiny{\textbf{\textcolor{gray}{(+1.47)}}}& 5.74 \tiny{\textbf{\textcolor{gray}{(+1.07)}}}& 5.24 \tiny{\textbf{\textcolor{gray}{(+0.57)}}}   \\
        \textbf{Average}        & 9.88  & 11.51 \tiny{\textbf{\textcolor{gray}{(+1.63)}}}& 10.94 \tiny{\textbf{\textcolor{gray}{(+1.06)}}}& 10.66 \tiny{\textbf{\textcolor{gray}{(+0.78)}}}   \\
        \bottomrule
    \end{tabular}

    \caption{Ablation experiments reveal the effectiveness of the Transformer, Graph Convolution, and global attention.}
    \label{table:abla_arch}
\end{table}

%% file: tables/ablation_graph_feature.tex
\begin{table}[t]
	\centering
	\footnotesize
	\setlength{\tabcolsep}{1.2mm}
    
	\begin{tabular}{l|cccccc}
		\toprule
		\multirow{3}{*}{} & \multicolumn{6}{c}{PCPNet Dataset}  \\
		\cline{2-7}
		& \multicolumn{4}{c}{Noise $\sigma$} & \multicolumn{2}{c}{Density}  \\
		& None & 0.12\% & 0.6\% & 1.2\% & Stripes & Gradient   \\
		\midrule
		$xyz$+$\Delta_{xyz}$+$\mathbf{f}$+$\Delta_{\mathbf{f}}$  & 3.99 & 8.97 & 15.85 & 20.98 & 4.81 & 4.67 \\
		$xyz$+$\Delta_{xyz}$+$\mathbf{f}$ 	& 4.74  & 9.23 & 16.24 & 21.76 & 5.85 & 5.45 \\
		$xyz$+$\mathbf{f}$+$\Delta_{\mathbf{f}}$      & 4.90  & 9.34 & 16.55 & 22.43 & 5.63 & 5.40 \\
		\bottomrule
	\end{tabular} 
	\caption{Ablation study on input features for Graph Convolution. $\Delta_{xyz}$ represents the difference in coordinates between a neighbor and the query point, while $\Delta_\mathbf{f}$ indicates the difference in features between a neighbor and the query point.}
    \label{table:abla_graph_conv_features}
\end{table}